\pdfoutput=1
\documentclass{article}

\usepackage{microtype}
\usepackage{graphicx}
\usepackage{booktabs}
\usepackage{amsmath}
\usepackage{amssymb}
\usepackage{hyperref}

\usepackage[preprint]{icml2026}
\usepackage{iftex}
\ifPDFTeX
  \usepackage[T1]{fontenc}
\fi
\ifXeTeX
  \usepackage{fontspec}
\fi

\icmltitlerunning{You Don't Need To Stay in The Loop}

\begin{document}

\twocolumn[
  \icmltitle{You Don't Need To Stay in The Loop:\\An Agentic Robotics Loop for Robot-Policy Improvement}

  \begin{icmlauthorlist}
    \icmlauthor{Hang Yu}{ind}
  \end{icmlauthorlist}
  \icmlaffiliation{ind}{Independent Researcher, United States}
  \icmlcorrespondingauthor{Hang Yu}{hangyu8123@gmail.com}
  \icmlkeywords{agentic robotics, vision-language-action models, autonomous experimentation, tool use, reproducibility}
  \vskip 0.3in
]

\printAffiliationsAndNotice{}

\begin{abstract}
Coding agents close the software loop for training a robot policy without having a human in the loop: a main agent manages the training, subagents analyze and execute, treat robot policies and hardware as tools and do the work. We port this architecture to robot-policy improvement, where one difference dominates the design: robotic tools such as trained policies, training pipelines, and data collection fail routinely, so a tool's quality must be measured, recorded at every call, and expired when the artifact behind it changes. AgenticRobotics is a backend-independent control plane in which an LLM controller drives disposable workers through durable train--evaluate--improve transactions: an immutable objective, controller-owned measurement, commit-keyed crash recovery, an evidence-graded skill library, and a tool registry with a standardized, recorded call surface. The title is an operational claim, not a selection claim: the operator can leave because promotion is evidence-gated, state is recoverable, and capability quality is derived from records, not because the loop picks better checkpoints than a human. On the one lineage we measured, it does not. The gates provide measured false-promotion control (0.001 false promotions per run with the hardened margin, against 0.005--0.021 with the shipped rule), anytime-valid decisions under optional stopping, zero lost or duplicate effects under kill injection, and a signed verifier that caught six of six artifact-tampering classes.
\end{abstract}

\section{Introduction}

Robot learning has strong inner-loop machinery, including generalist vision-language-action (VLA) policies, scalable imitation, RL post-training, and standardized suites such as LIBERO~\citep{libero}. However, the \emph{outer} research loop still relies on a person to interpret failures, pick the next lever, manage long jobs, compare noisy evaluations, and remember what did not work. ``Agentic robotics'' here therefore means an AI agent that conducts this outer loop of robot-policy improvement research, not the more common reading in which the robot itself acts agentically in the world~\citep{agentic_survey,embodied_agentic}.

\begin{figure*}[t]
  \centering
  \includegraphics[width=\textwidth]{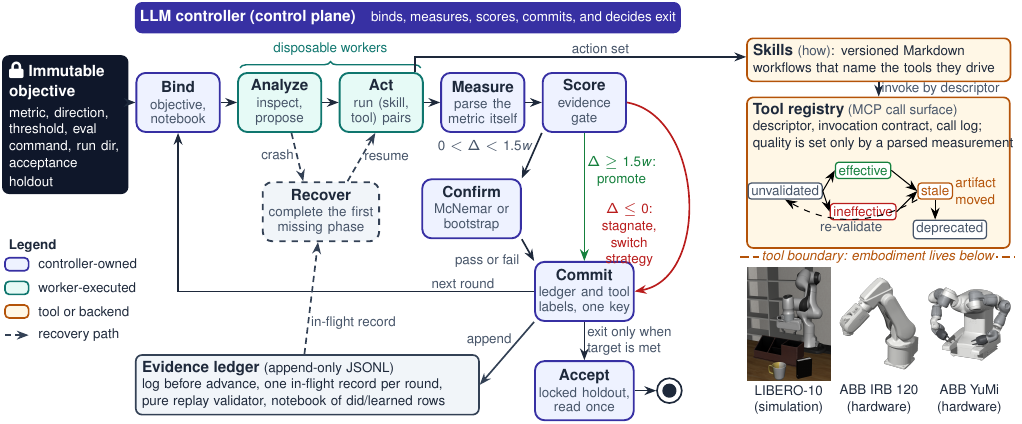}
  \caption{One AgenticRobotics round as a state machine (Section~\ref{sec:ideas}). The LLM controller owns Bind, Measure, Score, and Commit, and disposable workers execute Analyze and Act, which runs (skill, tool) pairs through the tool registry. Score is the evidence gate of Section~\ref{sec:gate}. It promotes a gain of at least $1.5w$, sends a smaller gain to paired confirmation, and counts no gain as stagnation without ending the run. Commit writes the ledger and the tool-quality labels under one commit key, and the run exits only when the target is met, through a locked acceptance holdout. After a crash, recovery completes the first missing phase from the in-flight record. Tool quality is set only by controller-parsed measurements and becomes stale when the artifact behind a tool changes. Robots stay below the tool boundary. The audited campaigns ran in LIBERO-10 simulation (rendered from the LIBERO environment), and the ABB IRB~120 and YuMi, rendered from their CAD models, illustrate hardware backends that would enter through the same descriptors. This report includes no hardware experiment.}
  \label{fig:pipeline}
\end{figure*}

This loop already has a working architecture in software engineering, and AgenticRobotics is built the way coding agents are built, on five design commitments (Figure~\ref{fig:pipeline}). \textbf{(i)}~The main agent manages the loop, and subagents analyze and execute. The controller spends its bounded context only on decisions, whereas disposable workers do the token-heavy work and return compact results. \textbf{(ii)}~Anything that executes in the world is a \emph{tool}, not knowledge, such as a trained policy, a training pipeline, a planner, or a data-collection rig. The controller never touches the robot's physics, since embodiment lives behind the tool boundary. \textbf{(iii)}~A packing workflow gives any such artifact one standardized input$\rightarrow$tool$\rightarrow$output call surface over the Model Context Protocol (MCP) and registers it, and a separate validating workflow measures what it can actually do. \textbf{(iv)}~Every call is recorded, and operational reliability is derived from records, never claimed. \textbf{(v)}~Skills are knowledgeable workflows for reaching the goal and for calling, building, validating, and improving tools. The load-bearing difference from software is that robotic tools are unreliable. A VLA policy measures mid-double-digit success, trainers crash, and hardware is sometimes absent, so tool-quality evaluation, data collection to improve a tool, and re-validation after retraining are first-class moves of the loop, themselves packaged as skills and tools.

The title is scoped by the evidence. ``Not staying in the loop'' is an \emph{operational} claim, which means that rounds proceed unattended because promotion is evidence-gated, state survives crashes, and capability trust expires with its artifact. It is not a claim of better selection than a human. On the one lineage we measured (Section~\ref{sec:experiments}), taking the final checkpoint beat every measurement-driven selector, ours included.

\textbf{Contributions.} \textbf{(1)}~A robotics research-round transaction with an immutable objective, controller-owned measurement, commit-keyed phases whose recovery completes rather than repeats, generation-scoped delegation, eleven invariants, and a pure replay validator. \textbf{(2)}~The agent$\rightarrow$skill$\rightarrow$tool decomposition, with a capability registry that binds measured competence to the artifact version that earned it and voids it on retraining, behind a standardized, per-call-recorded MCP call surface. None of the registry conventions surveyed in Section~\ref{sec:related} performs this binding. \textbf{(3)}~Measured operating characteristics of gate validity under optional stopping, crash recovery, signed measurement integrity, and fenced concurrency, together with a decision-quality study that includes where the loop loses. The hardened gate commits 0.001 false promotions per run, compared with 0.005--0.021 for the shipped rule, recovery loses or duplicates no effect over 14{,}000 kill injections, and the signed verifier detects six of six tamper classes. Every measurement-driven selector, ours included, loses 5.6--10.0 points of held-out success to taking the final checkpoint. \textbf{(4)}~Two audited campaigns (a screening reproduction and a failed improvement campaign that hardened the gate) and a positioning against concurrent systems (Table~\ref{tab:sota}).

\section{Related Work and the Gap}
\label{sec:related}

This section reviews five lines of prior work and, for each, states the gap that this work addresses.

\paragraph{Agentic optimization and audited agents.}
Voyager~\citep{voyager}, Eureka~\citep{eureka}, ADAS~\citep{adas}, AFlow~\citep{aflow}, and AlphaEvolve~\citep{alphaevolve} establish evaluator-grounded improvement of skills, rewards, agent programs, and workflows. Automated science extends the loop to entire papers~\citep{aiscientist}, and MLE-bench supplies the held-out evaluation convention that robot-policy loops still lack~\citep{mlebench}. Closest to our ledger, ASG-SI compiles an agent into an auditable skill graph with replayable promotion evidence~\citep{asgsi}, and execution provenance has been formalized as a typed graph over agent runs~\citep{agenttraces}. We specialize both to a robot-\emph{experiment} round whose commit keys, measurement generations, and promotion statistics survive process death, with measured operating characteristics rather than architecture alone.

\begin{table}[t]
\caption{Concurrent agent-driven robot-improvement systems: LearningFlow~\citep{learningflow}, RoboRouter~\citep{roborouter}, RHO~\citep{rho}, ENPIRE~\citep{enpire}, CaP-X~\citep{capx}, EvoTrainer~\citep{evotrainer}, ASPIRE~\citep{aspire}. Columns: durable evidence ledger; crash-recoverable transaction; measured, artifact-bound tool quality; released implementation. Entries follow the papers and project pages at the time of writing.}
\label{tab:sota}
\vskip -0.05in
\centering
\footnotesize
\setlength{\tabcolsep}{3pt}
\begin{tabular}{@{}lp{0.30\columnwidth}cccc@{}}
\toprule
System & Optimizes & Ledg. & Rec. & Meas. & Code \\
\midrule
LearningFlow & driving curricula, rewards & --- & --- & --- & --- \\
RoboRouter & policy selection & --- & --- & --- & --- \\
RHO & policy repository & --- & --- & --- & --- \\
ENPIRE & physical trials & --- & --- & --- & --- \\
CaP-X & code-as-policy skills & --- & --- & --- & \checkmark \\
EvoTrainer & training harness & --- & --- & --- & --- \\
ASPIRE & skill programs & --- & --- & --- & --- \\
\textbf{Ours} & experiment round & \checkmark & \checkmark & \checkmark & \checkmark \\
\bottomrule
\end{tabular}
\end{table}

\paragraph{Concurrent outer-loop systems.}
The systems in Table~\ref{tab:sota} differ in optimization unit, and none reports a durable evidence ledger, a crash-recoverable transaction, and measured tool quality together. CaP-X, with an open environment and benchmark, is the most tractable decision-quality comparator~\citep{capx}. EvoTrainer reaches promotion validity from the training-recipe side, reporting that evolving diagnostics prevent invalid high-scoring branches from promoting~\citep{evotrainer}. ASPIRE compounds a code-as-policy skill library but discloses no ledger or measured tool quality~\citep{aspire}.

\paragraph{Tool and capability registries.}
Agent frameworks standardize how a capability is \emph{described}: MCP, OpenAI function calling, and Claude tool use converge on a name, a what-and-when description, and a JSON-Schema input contract, with MCP's behavior annotations specified as untrusted hints~\citep{mcp_tools,openai_function_calling,anthropic_tool_use}. On the artifact side, Hugging Face's \texttt{model-index} separates what was measured from who verified it~\citep{hf_model_index}, and experiment trackers version metrics as lineage rows. None binds a capability's measured competence \emph{to the artifact version that earned it}, and no convention voids a recorded score when the weights behind a stable tool identity are retrained. RoboChallenge reports the same gap from the evaluation side, since it cannot verify that the model a user ran matches the model claimed~\citep{robochallenge}. Section~\ref{sec:capability} specifies that binding, for which we found no prior convention.

\paragraph{VLA backends.}
OpenVLA, $\pi_0$, $\pi_{0.5}$, SmolVLA, RT-2, Octo, OpenVLA-OFT, and SILVR define the policy layer we treat as a backend~\citep{openvla,pi0,pi05,smolvla,rt2,octo,openvlaoft,silvr}; interactive post-training, world-model RL, and residual-RL distillation move improvement inside the policy loop~\citep{interactive_vla,rise,pld}. PLD's near-saturated 99\% success on LIBERO indicates that a control plane's value cannot be read off a benchmark number that the backend already maximizes.

\paragraph{Evaluation and durable execution.}
LIBERO supplies the tasks~\citep{libero}, but robustness suites show that standard-split scores mask fragility~\citep{liberopro,liberoplus,liberox,realm}. Small-sample policy comparison demands interval and sequential statistics~\citep{precipice,step}. Our gate specializes anytime-valid inference, which includes N-SCORE for bounded metrics~\citep{beyond_binary}, optimal-policy identification~\citep{avopi}, betting-based sim-to-real certificates~\citep{simrealbetting}, and prediction-powered sim/hardware estimation~\citep{suresim}. Beldi and ExoFlow provide exactly-once patterns~\citep{beldi,exoflow}, and HELM addresses intra-episode execution memory~\citep{helm_exec}, which complements the inter-round level that this work addresses. Sim--real correlation studies~\citep{simpler} and online real-robot evaluation~\citep{robochallenge} motivate treating an evaluation backend and its validity domain as provenance.

\section{Control-Plane Design}
\label{sec:design}

This section first separates what the artifact implements from what a backend supplies, then states the four ideas that organize the design (Section~\ref{sec:ideas}), and then gives their mechanics: the controller and its workers, the round transaction, the promotion gate, and the capability registry.

\subsection{Artifact Boundary}

The \emph{protocol} is an executable Markdown controller loop that specifies eleven invariants, delegation rules, scoring, phase order, and exit verification. \emph{Reference code} implements the deterministic utilities (objective and command-preflight validation, a JSONL ledger, pure replay checking, and the tool registry with its call surface) and never executes training. The \emph{backend} supplies resumable training, checkpoints, simulators or robots, and an evaluation command that writes the target metric. The explicit boundary prevents a common ambiguity: a detailed protocol is not an implemented scheduler, and a utility is not evidence that every invariant held in every run.

\subsection{Four Design Ideas}
\label{sec:ideas}

Four ideas organize the architecture (Figure~\ref{fig:pipeline}), and the subsections that follow give their mechanics.

\paragraph{Toolify every execution.}
Nothing executes on the robot or its backend except through a tool. Each trained policy, training pipeline, evaluator, or teleoperated data-collection rig is packaged behind a descriptor. The descriptor states what the tool does, how it is invoked (argv template, JSON-Schema arguments, and required timeout), and what it has measurably done, and it keeps an append-only call history. Every tool is reachable through one standardized input$\rightarrow$tool$\rightarrow$output surface over MCP. There is no privileged code path around this boundary, so execution is \emph{enumerable} (the registry lists everything the system can do), \emph{auditable} (every call appends a record), and \emph{swappable} (a backend change is a new descriptor, not a new loop). The six seeded descriptors cover the audited campaigns' whole execution surface: policy, trainer, evaluator, recorder, dataset editor, calibrator.

\paragraph{Evaluate and label tools; never trust descriptions.}
A descriptor may say what a tool is \emph{for}; only measurement says what it can \emph{do}. Every tool carries per-benchmark quality on a five-status ladder (unvalidated, effective, ineffective, stale, and deprecated). This quality is settable only by a controller-parsed measurement, ranked by a Wilson lower bound so that a lucky small sample cannot outrank a well-measured mediocre one, and voided by derivation when the artifact behind the tool changes. Labeling is thus a continuing obligation of the loop, not a one-time act of registration: evaluating a tool, collecting data to improve it, and re-validating it after retraining are first-class round actions, themselves packaged as skills.

\paragraph{Skills are adjustable workflows over tools.}
A skill is an editable, versioned workflow, written as plain Markdown in the Agent Skills format, that says how to handle a situation by naming the tools it drives and the order and conditions under which to call them. Because the invocation contract lives in the tool descriptor rather than in the skill's prose, a skill can be adjusted (re-parameterized, reordered, or forked into a candidate variant) without touching execution code, and a tool's measured quality can change without editing a single skill. Skills therefore compound: new procedures enter the library as candidates, are exercised by rounds, and are promoted or deprecated on recorded evidence rather than authorial confidence.

\paragraph{Detach the physics.}
The controller never reasons about the robot as an embodied system. Motors, sensors, simulators, and rigs live entirely behind tool descriptors; what the control plane sees is a registry of evaluatable, callable capabilities with measured benchmark quality and derived operational reliability. This detachment makes the loop backend-independent. Swapping a simulator for hardware, or one robot for another, changes descriptors and their measurements, not the transaction, the gate, or a line of the protocol. The detachment also lets one uniform discipline govern different failure modes. A policy that succeeds 46\% of the time, a trainer that crashes, and an absent teleoperation rig are the same kind of object, which is a tool whose current label the recorded evidence does or does not support.

\subsection{Controller, Workers, Skills, and Tools}

The controller binds the objective, chooses actions, parses evaluation output itself, assigns verdicts, advances durable state, and decides exit. Workers perform bounded analysis, implementation, execution, QA, or adversarial verification (commitment~i). A generation identifier prevents a late worker or timer from waking a newer controller generation.

The action space decomposes as agent$\rightarrow$skill$\rightarrow$tool: the agent decides \emph{what}, a skill decides \emph{how}, and a tool is what the backend can actually \emph{do} (commitments~ii and~v). The skills directory is a menu, not a closed action space. At the audited snapshot it holds 29 loop skills and one vendored authoring tool, including four tool-lifecycle skills that build, pack, validate, and improve tools. Skills use the Agent Skills format~\citep{agentskills}. New procedures enter as candidates and advance only with evidence, because curated skills improve agent success while self-generated skills can degrade it~\citep{skillsbench}.

A tool is a capability with an artifact behind it, such as a trained policy, a training suite, a data collector, or an evaluator. Its descriptor states functionality, measured quality, the invocation contract, and an append-only history. The controller never invokes a tool directly, and a tool never decides anything. A skill names the tool it drives, and the executing worker takes the invocation contract from the descriptor rather than from prose. The indirection lets quality change without editing any skill and makes a backend swap one new descriptor. A round's action set is a set of (skill, tool) pairs.

For cross-run memory, a shared notebook holds one did/learned/source row per committed round. The controller reads the notebook at bind time and never reads it back into control state, and recurring lessons are distilled into the skill or tool they belong to~\citep{expel,generative_agents}. Each row is data, never instructions, so a command-shaped entry is a claim to be tested by this run's gates~\citep{memory_poisoning}.

The open action space is the practical difference from black-box optimizers such as Vizier~\citep{vizier}. There is no declared parameter space, since the controller may repair evaluation, inspect data, or adopt a retrieved method. This openness raises the burden of provenance and holdout discipline, because the search procedure itself adapts.

\subsection{Round Transaction}

An immutable objective specifies metric, direction, threshold, evaluation command, run directory, and optional acceptance holdout. Each round follows $\mathrm{Bind}\rightarrow\mathrm{Analyze}\rightarrow\mathrm{Act}\rightarrow\mathrm{Measure}\rightarrow\mathrm{Score}\rightarrow\mathrm{Commit}$. The controller stores one canonical in-flight record and advances side effects through commit-keyed phases under atomic replacement and a sidecar lock; recovery completes the first missing phase rather than replaying a committed effect. The protocol is inspired by exactly-once workflow systems~\citep{beldi,exoflow} but is not a proof of exactly-once execution, because bookkeeping correctness does not give backend idempotency. The intended contract is therefore a compound key \texttt{(round\_id, effect\_type, generation)} that the backend deduplicates, which Section~\ref{sec:experiments} tests and current training binaries do not implement. Tool registration uses the same phase and commit key, because two independently keyed phases would let a crash between them skip the tool update forever or double-apply an invalidation. Eleven invariants define success semantics: objective immutability; exit-only termination; controller-owned measurement; log-before-advance; an open menu of skills and tools; no runtime self-modification; generation-scoped wakeups; commit-keyed effects; placement independence; transitions derived only from verified work; and capability quality that is measured, never claimed, and never outlives its artifact.

\subsection{Evidence-Gated Promotion}
\label{sec:gate}

For binomial success counts $k_i/n_i$, the specification estimates a two-sample Agresti--Caffo half-width
\begin{equation}
\tilde p_i=\frac{k_i+1}{n_i+2},\quad
w=1.96\sqrt{\sum_{i=1}^{2}\frac{\tilde p_i(1-\tilde p_i)}{n_i+2}},
\end{equation}
and an operator may substitute a measured minimum delta. An improvement is promoted outright only when $\Delta=m_{\mathrm{new}}-m_{\mathrm{best}}\geq1.5w$ (sign reversed for minimization); a positive but smaller delta triggers confirmation, judged by McNemar or a paired bootstrap on paired episodes. Non-improvement increments stagnation, which forces a strategy-class switch but never terminates the run. An optional locked acceptance evaluation is read only at exit~\citep{reusable_holdout}. The interval is closed-form and stateless, so a context-limited controller need not persist accumulation state~\citep{precipice,beyond_binary}. Section~\ref{sec:experiments} measures the anytime-valid power that this choice forgoes. The SmolVLA campaign motivated the $1.5\times$ rule, since it promoted exactly at an 8-point band without confirmation.

\subsection{Capability Quality as Derived State}
\label{sec:capability}

The gate judges a checkpoint against the objective, whereas the registry answers what the controller is entitled to believe about a capability \emph{before} selecting one. The registry applies the five-status ladder of Section~\ref{sec:ideas} with four rules (commitment~iv). First, only a controller-parsed measurement may set effective or ineffective, and a worker's figure is provenance that leaves the tool unvalidated. Second, quality is keyed per benchmark, so concurrent runs cannot overwrite each other. Third, staleness is \emph{derived}, since each measurement records the artifact that produced it. When the tool's current artifact reference moves, which is exactly what retraining does, the tool reads stale automatically and reports ``not validated after retraining; previous rate $X$''. Fourth, quality may come only from a direct measurement of that tool, because a round's delta belongs to its whole action set. Candidates rank by the Wilson lower bound~\citep{wilson1927}, so a two-of-two fluke cannot outrank a well-measured 46-of-100. An unvalidated or stale tool cannot influence an unattended action set. This restriction is deliberately conservative, since it forces a revalidation that a score-carrying selector would skip.

\paragraph{The call surface.}
A tool that owns its command carries an invocation contract (argv template, JSON-Schema arguments, and a required timeout) and is callable through one pipeline (commitment~iii). The pipeline validates arguments, renders argv, checks an executable allowlist, uses no shell, and returns a result envelope in which launch failure and timeout are results, not exceptions. The same pipeline backs the CLI and a dependency-free MCP stdio server (2025-11-25 protocol revision; invocable tools only). Every call appends one record to an append-only log. \emph{Operational reliability}, whether the tool runs, is derived from those records as a Wilson lower bound and is deliberately distinct from benchmark quality, whether the artifact succeeds at the task. Three classes are deliberately not invocable: verbatim-evaluation commands (controller-owned measurement forbids delegating the metric), detached multi-hour trainers, and human-at-rig collection.

\paragraph{Status.}
The registry, invocation pipeline, and MCP server are implemented and unit-tested (106 tests) but have never executed in a logged round. The six seeded descriptors carry quality transcribed from evaluations reported here, no artifact reference has yet moved (so the staleness rule has never fired on a real trace), the invocation log holds no live-round calls, and the compounding benefit is untested (Section~\ref{sec:discussion}).

\section{Case Studies}
\label{sec:cases}

We audit ledgers and evaluation artifacts from two proof-of-concept campaigns on a Quadro RTX 8000. These are systems traces, not controlled comparisons of outer-loop algorithms.

\subsection{Task-5 VLA Screening}

The first campaign screened contemporary checkpoints and inference configurations from the $\pi_{0.5}$, MolmoAct2, GR00T~N1.7, and X-VLA families~\citep{pi05,molmoact2,groot,xvla} on one LIBERO-10 task (placing a book in a caddy). An $n{=}10$ probe selected this task for its highest baseline, so the results are task-conditional and invalid as a suite-level state of the art. Of ten retained configurations ($n{=}30$, fixed seed), seven measured 100\% and all met the campaign's 90\% gate. The fastest combined a larger action chunk with fewer flow-matching function evaluations (12.8 against 29.0 s/episode for the base $\pi_{0.5}$), adapting real-time chunking and flow distillation~\citep{rtc,snapflow} without claiming faithful reproduction. An incomplete two-task extension already bounds the result: GR00T and the chunk-50 $\pi_{0.5}$ each scored 83.3\% on Task~2. This is screening and reproduction, not a benchmark win. The full table is in the experiment package.

\subsection{SmolVLA Improvement Loop}

The second campaign targeted ${\ge}70\%$ mean success on the ten LIBERO-10 tasks with SmolVLA~\citep{smolvla}, warm-starting from a reported 55\% checkpoint at 100 episodes per evaluation. Rounds 0--26 committed over five calendar days (roughly three days of active GPU time). The operator stopped an incomplete round 27, so the loop never satisfied its own exit condition. A linear task-vector soup measured 63\% in round 2 and became the recorded champion. No later intervention beat it, since data rebalancing, more training, action-horizon changes, TIES merging, self-imitation, fresh retraining, and co-distillation all recorded 47--55\%. Repeated evaluation of that same champion pooled to 56.5\% over 1{,}200 episodes, so 63\% is a metric-best observation, not an estimate of skill. The historical $+8$-point promotion (McNemar $p\approx0.25$) would not pass the current 12-point outright threshold.

The trace shows a distinction that an autonomous loop must keep. A \emph{selection observation}, a \emph{champion record}, a \emph{performance estimate}, and an \emph{acceptance result} are four different quantities, and this campaign initially reported the first two as the third. The ledger did keep the loop from retrying closed intervention classes, which redirected effort toward data coverage. A separate from-scratch run reaches ${\approx}$46--50\%, which is consistent with community LIBERO-Long reproductions at 43--56\%, and serves as the multi-seed instrument of Section~\ref{sec:experiments}.

\section{Experiments: Operating Characteristics}
\label{sec:experiments}

This section measures how the gate, the transaction, and the measurement path behave, including where the loop loses. We ran every planned validation that was executable without a second trained lineage or a proprietary optimizer, and we measured three mechanisms that earlier revisions listed as future work, which are backend idempotency, signed measurement, and fenced fan-out. Artifacts are under \path{tech_report/experiments}, and unflattering results are reported as such.

\paragraph{Gate validity.}
A $2{\times}10^{5}$-replicate Monte-Carlo study at the campaign's own noise (twelve repeated reads of one champion: 56.5\%, SD 3.6 points) gives Table~\ref{tab:oc}. Under optional stopping, naive McNemar peeking inflates false promotion to 0.161 ($3\times$ the nominal $\alpha$). The fixed Agresti--Caffo $1.5\times$ rule is conservative (0.002) but underpowered at the historically relevant $+8$-point effect. Two anytime-valid tests, a testing-by-betting $e$-process (the binary specialization of N-SCORE~\citep{beyond_binary}) and a mixture-SPRT, hold $\alpha$ by construction via Ville's inequality and reach power 0.61--0.86 in fewer episodes on average. The closed-form gate is therefore safe but underpowered, and the anytime-valid upgrade recovers power under opportunistic stopping.

\begin{table}[t]
\caption{Promotion-gate operating characteristics under optional stopping at LIBERO noise ($2{\times}10^{5}$ Monte-Carlo replicates; null $p=0.565$, $n=100$ per look, decisions to 1{,}200 episodes). Type-I is the false-promotion rate over all looks; power and mean episodes-to-decision (ESS) at a true $+8$-point effect; dashes mark non-applicable entries.}
\label{tab:oc}
\centering
\footnotesize
\begin{tabular}{@{}lrrr@{}}
\toprule
Rule & Type-I & Power$_{+8}$ & ESS$_{+8}$ \\
\midrule
Agresti--Caffo $1.5\times$ (fixed) & 0.002 & --- & 100 \\
Min-delta $1.5\times$, 12\,pt (fixed) & 0.049 & 0.31 & 100 \\
McNemar, peeking & 0.161 & --- & --- \\
$e$-process (anytime-valid) & 0.003 & 0.61 & 920 \\
Mixture-SPRT (anytime-valid) & 0.007 & 0.86 & 700 \\
\bottomrule
\end{tabular}
\end{table}

\paragraph{Selection bias.}
Choosing the best of $K$ noisy 100-episode reads is $+9.8$ points optimistic at the campaign's $K{=}27$. A faithful replay of its 8-point rule reproduces the recorded 63\% champion at a true 56.5\%, while the hardened 12-point margin cuts the optimism to $+5.2$ points and roughly halves false promotions.

\paragraph{Recovery.}
Process kills at every ledger write boundary (14{,}000 injections), at all 78 byte offsets of a partial multi-byte append, and around each commit phase produced \emph{zero} lost and \emph{zero} duplicate side effects; recovery is idempotent and costs 0.15--0.25\,ms. This certifies the bookkeeping layer, not backend-command idempotency.

\paragraph{Multi-seed reality check.}
Re-evaluating the from-scratch run's champion (recorded 48\%) against its ``best observed'' checkpoint (recorded 53\%) at two fresh seeds, $n{=}100$ each, both measure 46.0\% (paired McNemar $p{=}1.0$, $b{=}c{=}33$). No difference was detected, which suggests that the recorded $+5$-point lead was selection noise, consistent with the winner's curse and with the gate's refusal to promote. Identical weights and seed also read 53\% and then 49\%, so fixed-seed rollouts are nondeterministic.

\paragraph{Decision quality, including where the loop loses.}
We measured a real surface of all eight checkpoints of the from-scratch lineage at two seeds, $n{=}100$ each (1{,}600 real LIBERO-10 episodes). A rule fixed before any score was read split the ten tasks 5/5 into a selection set and a locked holdout. A serial control re-measured one cell alone on the GPU. The aggregate was identical, but 22 of 100 episodes flipped outcome, so fixed-seed evaluation reproduces an episode's \emph{initial state}, not its outcome.

Table~\ref{tab:decision} is not the result the thesis would prefer. The oracle arm is the lineage's final checkpoint, so the trivial human script achieves zero regret at zero evaluation cost, while every measurement-driven strategy, ours included, loses 5.6--10.0 points of held-out success. The surface shows the mechanism. Selection and holdout tasks rank different checkpoints highest, so selecting on a measured score misleads, and more budget barely helps, since random search, TPE, and Thompson sampling are indistinguishable at every budget. The gates contribute the last column. The hardened $1.5\times$ margin commits $0.001$ false promotions per run, compared with $0.005$--$0.021$ for the shipped rule, a quantity that an argmax selector cannot express because it has no notion of a false promotion. Two caveats bound this result. A monotonically improving lineage is the easiest case for ``take the last checkpoint'' and the hardest for any selector, whereas a lineage that regresses late, as campaign~2 did, would be expected to invert the comparison. One lineage cannot separate these regimes.

\begin{table}[t]
\caption{Decision quality on the measured surface (budget 1{,}600 episodes; 2{,}000 bootstrap replicates; oracle 60.0\%). ``False prom.'' is promotions to an arm not truly better, defined only for sequential gates.}
\label{tab:decision}
\centering
\footnotesize
\begin{tabular}{@{}lrrr@{}}
\toprule
Strategy & Holdout & Regret & False prom. \\
\midrule
Human, last checkpoint & \textbf{60.00} & \textbf{0.00} & --- \\
Random search & 54.40 & 5.60 & --- \\
Vizier-style TPE & 54.33 & 5.67 & --- \\
Thompson sampling & 54.16 & 5.84 & --- \\
Human, best observed & 54.04 & 5.96 & --- \\
Ours, $e$-process & 53.94 & 6.06 & 0.067 \\
Ours, fixed 8-pt & 51.84 & 8.15 & 0.005 \\
Ours, hardened 12-pt & 50.00 & 10.00 & \textbf{0.001} \\
\bottomrule
\end{tabular}
\end{table}

\paragraph{Where the gate's validity comes from.}
Re-running the study under alternative noise models sharpens the answer. Pairing on identical initial states makes the per-evaluation latent effect \emph{shared} between champion and challenger, as the serial control confirms. With a shared effect, every anytime-valid rule holds Type-I at or below $\alpha$ under heavy overdispersion, the campaign's real per-task difficulty spread, and bounded partial-credit progress scores (0.0027--0.0032), while peeking inflates to ${\approx}0.185$ throughout. When the latent effect is \emph{arm-specific}, the per-episode martingale breaks and every per-episode rule inflates, the $e$-process included (0.118--0.505). The gate's validity rests on paired evaluation, not on the outcome distribution, so an unpaired multi-seed comparison must not reuse these guarantees. We also correct a defect in our earlier code: the betting fraction must satisfy $\lambda<2$ for outcomes in $[0,1]$ to keep the wealth process a nonnegative martingale.

\paragraph{Idempotency, integrity, and concurrency.}
(i)~A backend idempotency API, with keys scoped to \texttt{(run\_id, round\_id, effect\_type, generation)}, IETF-style replay/conflict semantics, and saga compensation~\citep{sagas}, passes an eight-property conformance harness. Five deliberately broken backends each fail a distinct property, and 4{,}000 crash injections between physical effect and durable receipt yield zero duplicate and zero lost effects. However, the actual LeRobot binaries carry no key and would re-run work. (ii)~A non-LLM parser that recomputes the metric from per-episode outcomes, plus an HMAC binding it to the checkpoint and command digest, detects six of six tamper classes over 27 real artifacts; the current parse path detects none. (iii)~Fenced fan-out, in which the aggregator rather than the lease service rejects stale-token partials~\citep{fencing}, survives 2{,}000 randomized schedules uncorrupted, while an unfenced aggregator is corrupted in 99.6\% of straggler trials.

\paragraph{Acceptance and cost.}
The locked acceptance holdout, previously specification-only, runs end to end: the holdout mechanically refuses reads outside the acceptance phase (three simulated unauthorized reads refused and logged), the accepted number is signed, and the gate returned \texttt{ACCEPTANCE\_FAILED} at 53.9\% against the 70\% target. The honest limit is instrument width: 100 held-out episodes give a $\pm$9.5-point interval against an 8-point \texttt{min\_delta}~\citep{suresim}. Read from artifacts, the cost was 121.6 GPU-hours over 5{,}475 real episodes, 9.9 of them for these experiments. Controller token cost was not retained, which is a design defect, since efficiency claims are unfalsifiable without it.

\section{Discussion}
\label{sec:discussion}

\paragraph{What the evidence supports, and what remains.}
The measured results support treating the outer loop as a device for \emph{controlling erroneous promotion}, through valid gates under optional stopping, zero lost or duplicate effects, tamper-evident measurement, and fenced aggregation. They do not support it as a device for finding better policies, since on the measured lineage its selections lost to the final checkpoint. Establishing more requires four further steps: the skill- and tool-compounding ablation (library and registry on versus off with live controllers, since the cheap hand-coded-prior version is circular); decision quality across campaigns including one that regresses late, against human scripts, Vizier~\citep{vizier}, and a runnable agent outer loop, of which CaP-X is the most tractable~\citep{capx}; generalization across suites and robustness perturbations~\citep{liberox,realm}; and a controller instrumented to log per-round token usage.

\paragraph{Limitations.}
This is a proof of concept. Two single-seed, small-$n$ campaigns on one benchmark suite motivate the design but do not validate it, and campaign~1 predates parts of the current transaction record, so the evidence package is not uniform. The restored objective schema is load-bearing, since the suite passes 106 tests and removing the file reintroduces nine failures. One defect is recorded rather than repaired: \path{tests/test_replay_properties.py} fails at collection when the optional \texttt{hypothesis} dependency is absent, despite being documented as auto-skipped. More broadly, language-model behavior and external services remain nondeterministic. Commit-keyed state prevents duplicate bookkeeping, not repeated physical consequences outside the declared backend contract.

\paragraph{Reproducibility.}
The report is a working-tree addition over base revision \href{https://github.com/HangYu8123/AgenticRobotics/tree/af89f02d8f886946b1870b75c166e514b75d7cc5}{\texttt{af89f02d8f88}}. The registry, call surface, and notebook are \path{tools/} (six descriptors), \path{agentic_robot/tools.py}, \path{agentic_robot/invocation.py}, \path{agentic_robot/mcp.py} with their tests, and \path{NOTEBOOK.md}. Every number in Sections~\ref{sec:cases} and~\ref{sec:experiments} is traceable to a script and result file in a named package under \path{tech_report/experiments}. The external LeRobot checkout is \href{https://github.com/huggingface/lerobot/tree/e40b58a8dfa9e7b86918c374791599d070518d11}{\texttt{e40b58a8dfa9}}; historical logs do not pin it. This revision's PDF was built with MiKTeX pdfLaTeX on Windows (ICML 2026 style archive vendored), and arXiv v1 with TeX Live pdfLaTeX on macOS. Earlier revisions used Tectonic 0.16.9.

\paragraph{Safety.}
Unbounded iteration is not permission for unbounded action. Objectives must constrain executables, paths, budgets, hardware access, and human approval points, and real-robot deployment requires independent safety interlocks, since a controller and a statistical gate are not safety certification.

\section{Conclusion}

You do not need to stay in the loop, in a precise and deliberately narrow sense. A campaign runs unattended because each of four dangerous decisions is governed: promotion by an evidence gate that stays valid under opportunistic stopping, interruption by a commit-keyed transaction that recovers without duplicating work, capability trust by a registry whose measured quality expires with its artifact, and measurement by a controller that parses the metric itself. The gates produced fewer false promotions, not better selections, so leaving the loop rests on control of error rather than on better-than-human choice. The comparisons listed in Section~\ref{sec:discussion} would test whether the loop can also select better than a human.

\bibliography{references}
\bibliographystyle{icml2026}

\end{document}